\documentclass[runningheads]{llncs}
\usepackage{graphicx}
\usepackage{amsmath,amssymb} % define this before the line numbering.
\usepackage{color}
\usepackage{subfigure}
\begin{document}
\title{Hyperspectral City V1.0\\ Dataset and Benchmark} 
% Replace with your title

\titlerunning{Hyperspectral City\\ Dataset and Benchmark}
% Replace with a meaningful short version of your title
%
\author{Shaodi You \inst{1} \and
Erqi Huang \inst{2} \and
Shuaizhe Liang \inst{3} \and
Yongrong Zheng \inst{3} \and
Yunxiang Li \inst{4} \and
Fan Wang \inst{5} \and
Sen Lin \inst{2} \and
Qiu Sheng \inst{2} \and
Xun Cao \inst{2} \and
Diming Zhang \inst{2}$^,$\inst{11} \and
Yuanjiang Li \inst{11} \and
Yu Li \inst{6} \and
Ying Fu \inst{3} \and
Boxin Shi \inst{7} \and
Feng Lu \inst{8} \and
Yinqiang Zheng \inst{9} \and
Robby~T.~Tan \inst{10}
}

%
%Please write out author names in full in the paper, i.e. full given and family names. 
%If any authors have names that can be parsed into FirstName LastName in multiple ways, please include the correct parsing, in a comment to the volume editors:
%\index{Lastnames, Firstnames}
%(Do not uncomment it, because you may introduce extra index items if you do that, we will use scripts for introducing index entries...)
\authorrunning{Shaodi You}
% Replace with shorter version of the author list. If there are more authors than fits a line, please use A. Author et al.
%

\institute{Corresponding Author, University of Amsterdam, Netherlands
\and Nanjing University, China
\and Beijing Institute of Technology, China
\and ZVision Co. Ltd. China
\and Zongmu Co. Ltd. China
\and Tencent, China
\and Peking University, China
\and Beihang University, China
\and National Institute of Informatics, Japan
\and National University of Singapore, Singapore
\and Jiangsu University of Science and Technology, China
}
\maketitle              % typeset the header of the contribution
\begin{abstract}
This document introduces the background and the usage of the Hyperspectral City Dataset and the benchmark. The documentation first starts with the background and motivation of the dataset. Follow it, we briefly describe the method of collecting the dataset and the processing method from raw dataset to the final release dataset, specifically, the version 1.0. 
We also provide the detailed usage of the dataset and the evaluation metric for submitted the result for the 2019 Hyperspectral City Challenge.

\end{abstract}
\section{Background and Motivation}

\textbf{Physics Based Vision Meets Deep Learning}

Light traveling in the 3D world interacts with the scene through intricate processes before being captured by a camera. These processes result in the dazzling effects like color and shading, complex surface and material appearance, different weathering, just to name a few. Physics based vision aims to invert the processes to recover the scene properties, such as shape, reflectance, light distribution, medium properties, etc., from the images by modeling and analysing the imaging process to extract desired features or information. 

There are many popular topics in physics based vision. Some examples are shape from shading, photometric stereo, reflectance modelling, reflection separation, radiometric calibration, intrinsic image decomposition, and so on. As a series of classic and fundamental problems in computer vision, physics based vision facilitates high-level computer vision problems from various aspects. For example, the estimated surface normal is a useful cue for 3D scene understanding; the specular-free image could significantly increase the accuracy of image recognition problem; the intrinsic images reflecting inherent properties of the objects in the scene substantially benefit other computer vision algorithms, such as segmentation, recognition; reflectance analysis serves as the fundamental support for material classification; and, bad weather visibility enhancement is important for outdoor vision systems.

In recent years, deep neural networks and learning techniques show promising improvement for various high-level vision tasks, such as detection, classification, tracking, etc. With the physics imaging formation model involved, successful examples can also be found in various physics based vision problems (please refer to the references section). 

When physics based vision meets deep learning, there will be mutual benefits. On one hand, classic physics based vision tasks can be implemented in a data-fashion way to handle complex scenes. This is because, a physically more accurate optical model can be too complex as an inverse problem for computer vision algorithms (usually too many unknown parameters in one model), however, it can be well approximated providing a sufficient collection of data. Later, the intrinsic physical properties are likely to be learned through a deep neural network model. Existing research has already exploited such benefit on luminance transfer, computational stereo, haze removal, etc.

On the other hand, high-level vision task can also be benefited by awareness of the physics principles. For instance, physics principles can be utilized to supervise the learning process, by explicitly extracting the low-level physical principles rather than learning it implicitly. In this way, the network could be more accurate more efficient. Such physics principles have already presented the benefits in semantic segmentation, object detection, etc.
Therefore, we believe when physics based vision meets deep learning both low level and high level vision task can get the benefits. Furthermore, we believe that there are many computer vision tasks that can be tackled by solving both physics based vision and high level vision in a joint fashion to get more robust and accurate results which cannot be achieved by ignoring each side. 

\noindent\textbf{Hyperspectral City}

We propose a semantic segmentation challenge for urban autonomous driving scene which utilizes newly developed hyperspectral camera. The motivation is to compensate the insufficient visual quality problem of existing dataset. Particularly, the CityScape \cite{cordts2016cityscapes} dataset provides only extremely washed out RGB images. To solve this, we endeavour to propose the new dataset which adopts multi-channel visual input. 
Our new dataset, can provide the following benefits: 1. properly balanced and colourful visual input. 2. We can analyse and see visual properties which cannot be seen from RGB channels. 3. We can robustly handle night scenes, thanks to the near infrared band. 4. We can robustly handle water phenomenon including rain and fog, because of the absolution behaviour in the infrared band. 

For the initial release of the dataset, we decide to propose the task of  semantic segmentation with coarse labeling. 
We release 367 frames hyperspectral images with coarse labeling for training and 55 frames with fine labeling for testing.

\begin{figure}[tbp]%
  \begin{center}
    \includegraphics[width= 0.9 \linewidth]{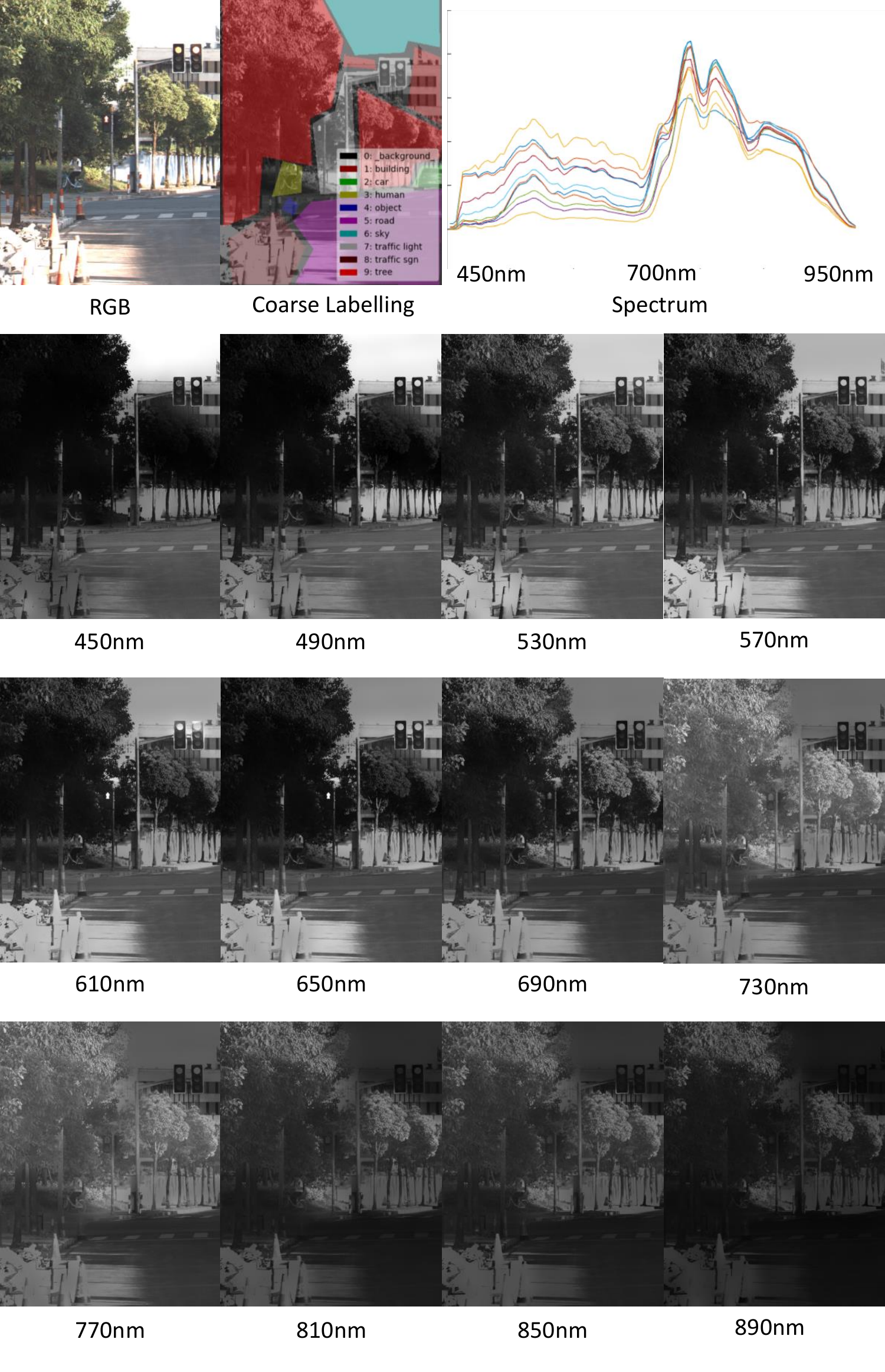}
  \end{center}
  \caption{An example of the Hyperspectral City 1.0 Dataset, we provide high resolution hyperspectral data in typical driving scenes with coarse semantic label.}
  \label{fig:HSD}
\end{figure}

\begin{figure}[t]%
  \begin{center}
    \includegraphics[width=\linewidth]{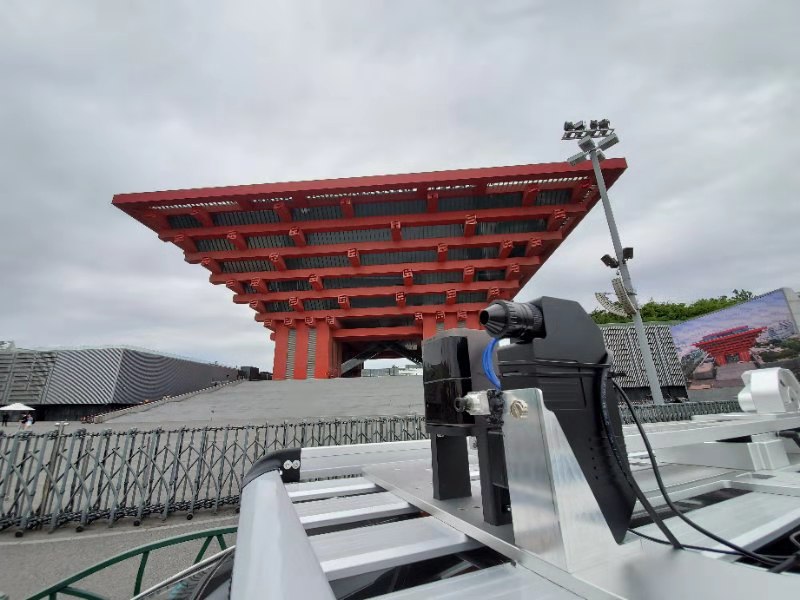}
  \end{center}
  \caption{Outdoor data capturing in Shanghai. The camera system is top-mounted.}
  \label{fig:shanghai}
\end{figure}

%%%%%%%%%%%%%%%%%%%%%%%%%%%%%%%%%%%%%%%%%%%%%%%%%%%%%%%%%
%SECTION DATASET GENERATION
%%%%%%%%%%%%%%%%%%%%%%%%%%%%%%%%%%%%%%%%%%%%%%%%%%%%%%%%%

\section{Dataset Generation}

\subsection{Data Collection}

\begin{figure}[t]%
  \begin{center}
    \includegraphics[width=\linewidth]{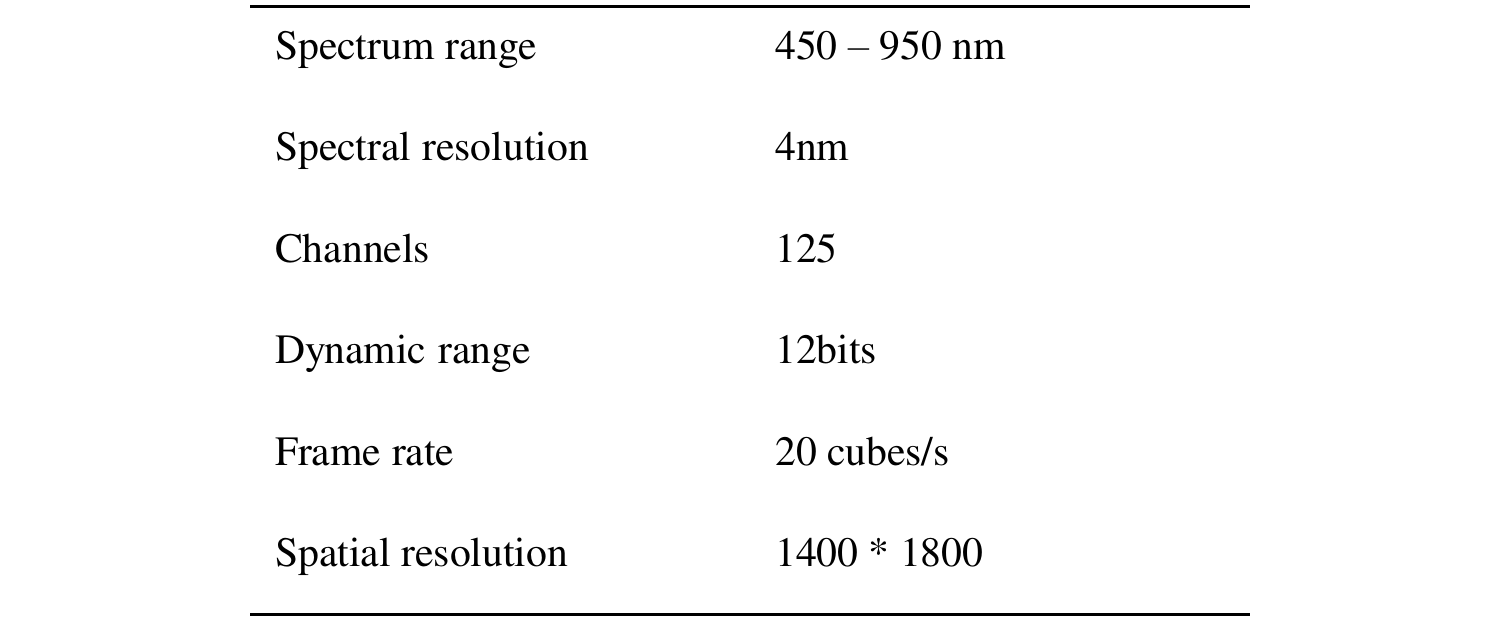}
  \end{center}
  \caption{Specifications of the hyperspectral data}
  \label{fig:lightgene}
\end{figure}

\paragraph{Hyperspectral Sensor}
We use the LightGene Hyperspectral Sensor for the data collect. Fig. \ref{fig:lightgene} is a brief review of the LightGene camera sensor. In particular, the camera can provide hyperspectral dataset in the range of 450 to 950nm with a spectral resolution at 4nm. In total, the camera can provide 125 spectral channels. The spectral resolution of each channel is approximately 1400 by 1800 pixels. And therefore, in total, each frame of the hyperspectral image is with the size:
\begin{equation}
1400 * 1800 * 125.
\label{eq:size}
\end{equation}

\paragraph{Outdoor data collection in Shanghai}

The dataset collection is in Shanghai for three days in June. The we collected data in a variety of environment including: crowded traffic area, famous buildings and structures, CBD, highways, quiet suburbs, overpasses and underground parking. The weather condition includes sunny and cloudy days. And the lighting condition includes day, night and sunset. We use standard color board for color calibration. We collecting data, the car is driving at a speed in the range of 20-50km/h. The hyperspectral camera is working at 1fps. The field of view (FOV) is 9 degrees in current lens configuration. The camera is vertical mounted to enable capturing of a wider dynamic range.

\begin{figure}[t]%
  \begin{center}
    \includegraphics[width=\linewidth]{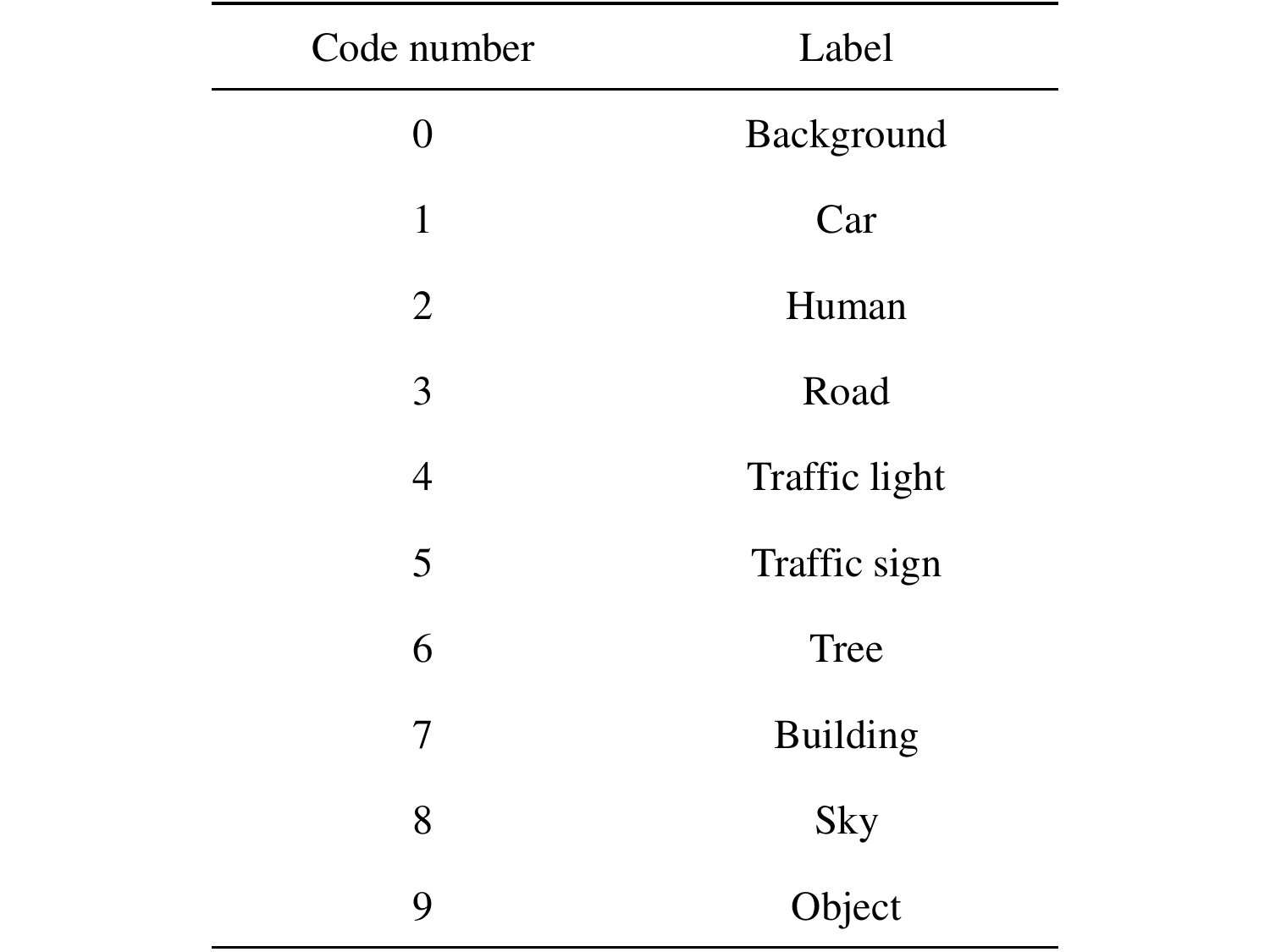}
  \end{center}
  \caption{A summary of dataset label}
  \label{fig:label}
\end{figure}

\subsection{Dataset Labeling}

\paragraph{Data selection}
The V1.0 dataset is focusing on semantic segmentation using coarse labeling. It is aiming to exploit the rich information in the hyperspectral data. Therefore, we manually selected 367(training) plus 55(testing) hyperspectral images which are considered suitable for the semantic segmentation task.

\paragraph{Coarse Labeling}
For the training dataset, we provide only the coarse labeling. The labeling is deliberate conducted in a quick and rough fashion from 10 different people. And the level of detail is various from people to people. Therefore, we are encouraging the users to learn information from the rich hyperspectral information rather than from human labeling.
Fig.\ref{fig:HSD} shows an example of the coarse labeling. We label 300 images from the dataset and consider them as training set.

We use LabelMe \cite{russell2008labelme} as the labeling tool. And the labels and colors are consistent with CitiScape\cite{cordts2016cityscapes} dataset.

\paragraph{Fine-grained Labeling}
For testing purpose, we also fine-grained labeled 55 images. During the challenge, only the input hyperspectral images are available and the groundtruth will not be released.

\subsection{Hyperspectral Image Compression}
Unlike RGB images which only has 3 channels, each hyperspectral images in our dataset has 125 channels. Without compression, each hyperspectral cubic will take more than 1Gigabytes storage and will make it unreasonable for online transfer. In our current release, we compress the image using H.264 encoder and decoder with a quality setting at 90\%. The compression ratio is smaller than 2\% and the final dataset size is around 40 Gigabytes.

\section{Dataset Usage}

\subsection{Access the Challenge Dataset}

Challenge dataset can be downloaded from the cloud platform, please visit the challenge website for detail:

\begin{center}
    \url{https://pbdl2019.github.io/index.html}.
\end{center}

\subsection{Content of The Downloaded Package}

In the dataset, you may find a folder for training data and a folder for testing data. 

\paragraph{train} 
In the training dataset folder, you will find 367 sub-folders, each of which is named with an index. In each of the folder, you will find five files, which are:
\begin{equation}
\nonumber
<*.hsd>, <*\_cropped.png>, <label\_gray.png>, <label\_vis.png>,
\label{eq:train}
\end{equation}
They are: 1. the hyperspectral data file, 2. The RGB visualization of the hyperspectral image, 3. Semantic labelling coded from 0 to 9 as per Fig.\ref{fig:label}, 4. Visualization of the semantic labelling by overlay with the RGB image. The $*$ is the filename which is the same as the folder name.

\paragraph{test}
In the testing dataset, you will find 68 sub-folders. The same as training dataset, each of the sub-folder is named with a file index. In each of the folder, you will find two files, which are:
\begin{equation}
\nonumber
<*.hsd>, <*\_cropped.png>.
\label{eq:test}
\end{equation}
They are: 1. the hyperspectral data file, 2. The RGB visualization of the hyperspectral image. The $*$ is the filename which is the same as the folder name.

\paragraph{Read Hyperspectral Cubic Dataset}
In the dataset package, you will find a MATLAB file named \textit{readHSD.m}, which is the MATLAB code to read the hyperspectral data. The data format is similar as RGB images, the different is the file has 125 channels rather than 3 channels.

\section{Evaluation Metric}

In the challenge, we use mean Intersection over Union as the evaluation metric.

\subsection{Website for submission}
You may submitted your result to the following website.

\subsection{Format of Submission}
To submit your result, you are required to submit a $.zip$ file. You may pickup any name for the zip file.

\paragraph{Format of Filenames}
Within the zip file, you are required to name each of the segmentation result using the same name as the folder name. For example, the segmentation result for folder named:
\begin{equation}
\nonumber
rgb20190528\_180044\_6684\_json
\end{equation}
should be
\begin{equation}
\nonumber
<rgb20190528\_180044\_6684\_json.png>
\end{equation}
All the png files should be under the same folder, there should be no subfolders in your submission.

\paragraph{Format of Images}

You are required to submit you segmentation result using the specified code of color. The segmentation result should be stored as png image. Specifically, 8-bit, single channel, L mode png format.
The color code should be the same as provided in Fig. \ref{fig:label}.

\vspace{24pt}
\begin{large}\noindent\textbf{Acknowledgement}\end{large}
\vspace{12pt}
We thank the following people in organizing the challenge.
 Dr. Yu Li, Prof. Yin Fu, Mr. Shuangzhe Liang, Mr. Yongrong Zheng.

We thank ZONGMU Co. Ltd. to facilitate with the vehicle set up and providing the car fleet for data collection.

We thank Prof. Xun Cao, Mr. Sen Lin, Dr. Qiu Shen, Mr. Erqi Huang from  Nanjing University in providing the LightGene hyperspectral camera and facilitate the data collection.

We thank Dr. Yunxiang Li from ZVISION Technologies Co., Ltd. in providing the Solid State Laser Scanner and facilitate the data collection.

We thank the following people in contributing to the dataset labeling. Dr. Diming Zhang, A/Prof. Yuanjiang Li.

We thank the following people in organizing the 2nd ICCV Joint Workshop on Physcis Based Vision meets Deep Learning: Dr. Yu Li, Prof. Ying Fu, Dr. Shaodi You, A/Prof. Yinqiang Zheng, Prof. Feng Lu, Prof. Boxin Shi and Prof. Robby T. Tan.

\bibliographystyle{splncs04}
\bibliography{reference}

\end{document}